\documentclass[10pt,twocolumn,letterpaper]{article}

\usepackage{iccv}
\usepackage{times}
\usepackage{epsfig}
\usepackage{graphicx}
\usepackage{amsmath}
\usepackage{amssymb}
\usepackage{tabularx, booktabs, ragged2e}
\usepackage[]{algorithm2e}
\usepackage{multirow}
\usepackage[acronym]{glossaries}
\usepackage{balance}

\newacronym{aflw}{AFLW}{Annotated Facial Landmarks in the Wild}
\newacronym{nmse}{NMSE}{Normalized Mean Square Error}
\newacronym{gan}{GAN}{generative adversarial network}
\newacronym{d}{$D$}{discriminator}
\newacronym{g}{$G$}{generator}
\newcommand{\image}{\mathbf{d}}

\newcommand{\hm}{\mathbf{h}}

\newcommand{\lmk}{{\mathbf{s}}}

\newcommand{\lkl}{\textrm{LaplaceKL}}
\newcommand{\sam}{\textrm{softargmax}}
\newcommand{\Sam}{\textrm{Softargmax}}

\usepackage[pagebackref=true,breaklinks=true,letterpaper=true,colorlinks,bookmarks=false]{hyperref}

\iccvfinalcopy 

\def\iccvPaperID{3137} 

\ificcvfinal\fi
\graphicspath{{figures/}} 

\begin{document}

\title{Laplace Landmark Localization}

\author{Joseph P Robinson$^1$\thanks{The work was done while the first author was interning at Snap Inc.} , Yuncheng Li$^{2}$, Ning Zhang$^{2}$, Yun Fu$^{1}$, and Sergey Tulyakov$^{2}$\\
\hspace{-.4in}$^{1}$Northeastern University\hspace{.7in} $^{2}$Snap Inc. \\\vspace{2mm}
} 

\maketitle
\ificcvfinal\thispagestyle{empty}\fi

\begin{abstract}
Landmark localization in images and videos is a classic problem solved in various ways. Nowadays, with deep networks prevailing throughout machine learning, there are revamped interests in pushing facial landmark detectors to handle more challenging data. Most efforts use network objectives based on $L_1$ or $L_2$ norms, which have several disadvantages. First of all, the generated heatmaps translate to the locations of landmarks (\ie confidence maps) from which predicted landmark locations (\ie the means) get penalized without accounting for the spread: a high-scatter corresponds to low confidence and vice-versa. For this, we introduce a $\lkl$ objective that penalizes for
low confidence. Another issue is a dependency on labeled data, which are expensive to obtain and susceptible to error. To address both issues, we propose an adversarial training framework that leverages unlabeled data to improve model performance. Our method claims state-of-the-art on all of the 300W benchmarks and ranks second-to-best on the \Gls{aflw} dataset. 
Furthermore, our model is robust with a reduced size: $1/8$ the number of channels (\ie 0.0398 MB) is comparable to the state-of-the-art in real-time on CPU. Thus, this work is of high practical value to real-life application.

\end{abstract}

\section{Introduction}
\label{sec:introduction}
\glsreset{aflw}
To localize landmarks is to find pixel locations in visual media corresponding to points of interest. In face alignment, these points correspond to face parts. For bodies and hands, landmarks correspond to projections of joints on to the camera plane~\cite{supancic2015depth, wang2019ev}. Historically, landmark detection and shape analysis tasks date back decades: from Active Shape Models~\cite{cootes1992active} to Active Appearance Models~\cite{cootes2001active}, with the latter proposed to analyze and detect facial landmarks.

\begin{figure}
    \centering
    \includegraphics[width=1.0\linewidth]{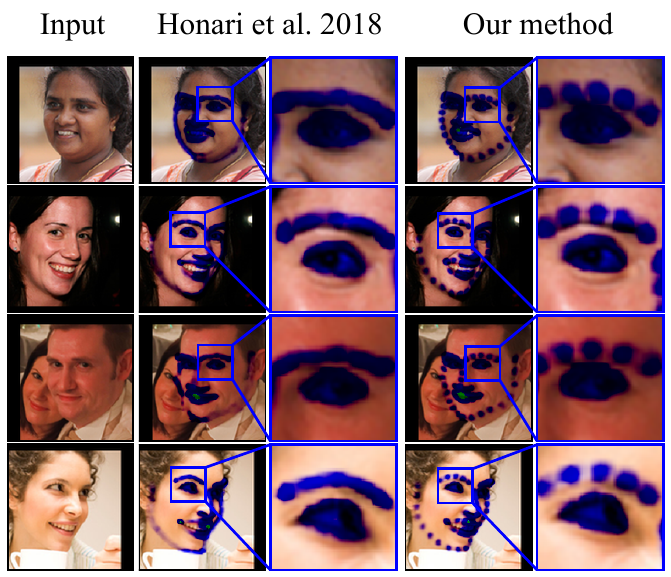}
    \caption{Heatmaps generated by $\sam$-based models (middle block) and the proposed $\lkl$ (right block), each with heatmaps on the input images (left) and a zoomed-in view of an eye region (right). These heatmaps are confidence scores (\ie probabilities) that a pixel is a landmark. softargmax-based methods generate highly scattered mappings (low certainty), while the same network trained with our loss is concentrated (\ie high certainty). We further validate the importance of minimizing scatter experimentally (Table~\ref{tab:face-comparisons}). Best if viewed electronically.} 
    \label{fig:kl-versus-softargmax}
\end{figure}

A need for more advanced models to handle increasingly tricky views has triggered revamped interest in facial landmark localization. Thus, came a wave of different types of deep neural architectures that pushed state-of-the-art on more challenging datasets. These modern-day networks are trained end-to-end on paired labeled data $(\image, \lmk)$, where $\image$ is the image and $\lmk$ are the actual landmark coordinates. Many of these used encoder-decoder style networks to generate feature maps (\ie heatmaps) to transform into pixel coordinates~\cite{newell2016stacked,peng2018red, yang2017stacked}. The network must be entirely differentiable to train end-to-end. Hence, the layer (or operation) for transforming the $K$ heatmaps to pixel coordinates must be differentiable~\cite{honari2018improving}. Note that each of the $K$ heatmaps corresponds to the coordinates of a landmark. Typically, the $\sam$ operation determines the location of a landmark as the expectation over the generated 2D heatmaps. Thus, metrics like or determine the distance between the actual and predicted coordinates $\tilde{\lmk}$, \ie $\mathbf{e}=\tilde{\lmk} - \lmk$.

There are two critical shortcomings of the methodology discussed above. (1) These losses only penalize for differences in mean values in coordinate space, and with no explicit penalty for the variance of heatmaps. Thus, the generated heatmaps are highly scattered: high variance means low confidence. (2) This family of objectives is entirely dependent on paired training samples (\ie $(\image, \lmk)$). However, obtaining high-quality data for this is expensive and challenging. Not only does each sample require several marks, but unintentional, and often unavoidable, labels are of pixel-level marks subject to human error (\ie inaccurate and imprecise ground-truth labels). All the while, plenty of unlabeled face data are available for free.

In this paper, we propose a practical framework to satisfy the two shortcomings. Thus, our first contribution alleviates the first issue. For this, we introduce a new loss function that penalizes for the difference in distribution defined by location and scatter (Fig.~\ref{fig:kl-versus-softargmax}). Independently, we treat landmarks as random variables with $\textrm{Laplace}(\lmk, 1)$ distributions, from which the KL-divergence between the predicted and ground-truth distributions defines the loss. Hence, the goal is to match distributions, parameterized by both a mean and variance, to yield heatmaps of less scatter (\ie higher confidence). We call this objective the LaplaceKL loss.

Our second contribution is an adversarial training framework for landmark localization. We propose this to tackle the problem of paired data requirements by leveraging unlabeled data for free. We treat our landmark detection network as a \gls{g} of normalized heatmaps (\ie probability maps) that pass to the \gls{d} to learn to distinguish between the true and generated heatmaps. This allows for large amounts of unlabeled data to further boost the performance of our $\lkl$-based models. In the end, \gls{d} proves to improve the predictive power of the $\lkl$-based model by injecting unlabeled data into the pipeline during training. As supported by experiments, the adversarial training framework complements the proposed $\lkl$ loss (\ie an increase in unlabeled data results in a decrease in error). To demonstrate this, we first show the effectiveness of the proposed loss by claiming state-of-the-art without adversarial training– to then further improve with more unlabeled data added during training!

Furthermore, we reduced the size of the model by using $\frac{1}{16}$, $\frac{1}{8}$, $\frac{1}{4}$, and $\frac{1}{2}$ the original number of convolution filters, with the smallest costing only 79 Kb on disk. We show an accuracy drop for models trained with the proposed $\lkl$ as far less than the others trained with a $\sam$-based loss. So again, it is the case that more unlabeled training data results in less of a performance drop at reduced sizes. It is essential to highlight that variants of our model at or of larger size than $1/8$ the original size compare well to the existing state-of-the-art. We claim that the proposed contributions are instrumental for landmark detection models used in real-time production, mobile devices, and other practical purposes.

Our contributions are three-fold: (1) A novel Laplace KL-divergence objective to train landmark localization models that are more certain about predictions; (2) An adversarial training framework that leverages large amounts of unlabeled data during training; (3) Experiments that show our model outperforms recent works in face landmark detection, along with ablation studies that, most notably, reveal our model compares well to state-of-the-art at $1/8$ its original size (\ie $<$160 Kb) and in real-time (\ie$>$20 fps).

\section{Related work}
\label{sec:relatedwork}

In this section, we review relevant works on landmark localization and \gls{gan}.

\textbf{Landmark localization} has been of interest to researchers for decades. At first, most methods were based on Active Shape Models~\cite{cootes1992active} and Active Appearance Models~\cite{cootes2001active}. Then, Cascaded Regression Methods (CRMs) were introduced, which operate sequentially; starting with the average shape, then incrementally shifting the shape closer to the target shape. CRMs offer high speed and accuracy (\ie $>$1,000 fps on CPU~\cite{ren2014face,kazemi2014one}). 

More recently, deep-learning-based approaches have prevailed in the community due to end-to-end learning and improved accuracy. Initial works mimicked the iterative nature of cascaded methods using recurrent convolutional neural networks~\cite{peng2018red, trigeorgis2016mnemonic,wang2016recurrentaccv,wang2018recurrentpami}. Besides, several have been several methods for dense landmark localization~\cite{guler2017densereg,jeni2015dense} and 3D face alignment~\cite{tulyakov2018consistent, zhu2016face} proposed: all of which are fully-supervised and, thus, require labels for each image. 

\begin{figure*}
    \centering
    \includegraphics[width=\linewidth]{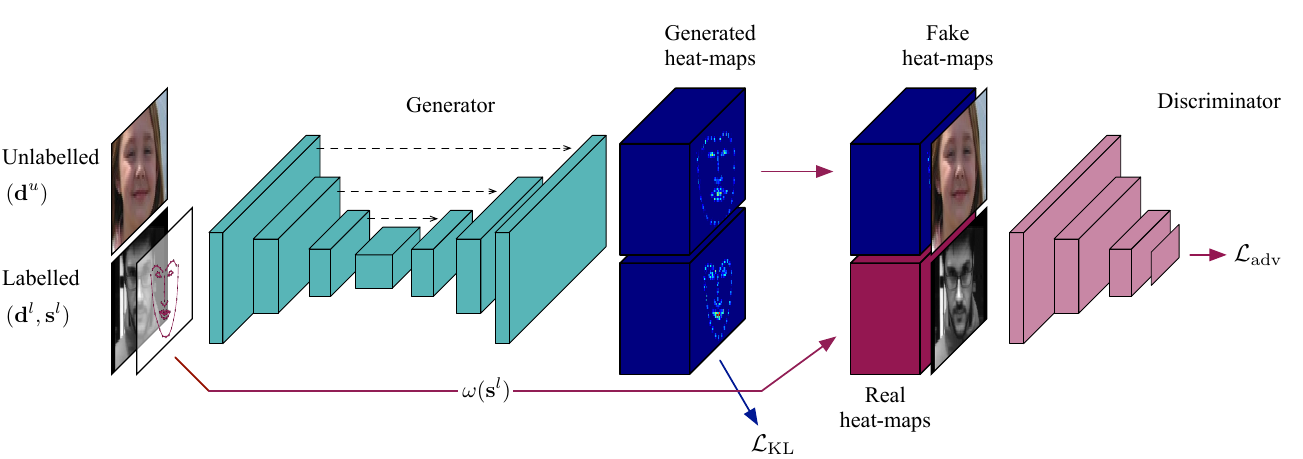}
    \caption{The proposed semi-supervised framework for landmarks localization. The labeled and unlabeled branched are marked with \textcolor{blue}{blue} and \textcolor{red}{red} arrows, respectfully. Given an input image, \gls{g} produces $K$ heatmaps, one for each landmark. Labels are used to generate real heatmaps as~$\omega(\lmk^l)$. \gls{g} produces fake samples from the unlabeled data. Source images are concatenated on heatmaps and passed to \gls{d}.}
    \label{fig:framework}
\end{figure*}

Nowadays, there is an increasing interest in semi-supervised methods for landmark localization. Recent work used a sequential multitasking method which was capable of injecting labels of two types into the training pipeline, with one type constituting the annotated landmarks and the other type consisting of facial expressions (or hand-gestures)~\cite{honari2018improving}. The authors argued that the latter label type was more easily obtainable, and showed the benefits of using both types of annotations by claiming state-of-the-art on several tasks. Additionally, they explore other semi-supervised techniques (\eg equivariance loss). In~\cite{dong2018supervision}, a supervision-by-registration method was proposed, which significantly utilized unlabeled videos for training a landmark detector. The fundamental assumption was that the neighboring frames of the detected landmarks should be consistent with the optical flow computed between the frames. This approach demonstrated a more stable detector for videos, and improved accuracy on public benchmarks. 

Landmark localization data resources have significantly evolved as well, with the 68-point mark-up scheme of the MultiPIE dataset~\cite{gross2010multi} widely adopted. Despite the initial excitement for MultiPIE throughout the landmark localization community~\cite{zhu2012face}, it is now considered one of the easy datasets captured entirely in a controlled lab setting. A more challenging dataset, \gls{aflw}~\cite{koestinger2011annotated}, was then released with up to 21 facial landmarks per face (\ie occluded or ``invisible'' landmarks were not marked). Finally, came the 300W dataset made-up of face images from the internet, labeled with the same 68-point mark-up scheme as MultiPIE, and promoted as a data challenge~\cite{sagonas2013300}. Currently, 300W is among the most widely used benchmarks for facial landmark localization. In addition to 2D datasets, the community created several datasets annotated with 3D keypoints~\cite{bulat2017far}.

\textbf{\Glspl{gan}} were recently introduced~\cite{goodfellow2014generative}, quickly becoming popular in research and practice. \Glspl{gan} have been used to generate images~\cite{radford2015unsupervised} and videos~\cite{saito2017temporal,tulyakov2017mocogan}, and to do image manipulation~\cite{geng20193d}, text-to-image\cite{han2017stackgan}, image-to-image~\cite{zhu2017unpaired}, video-to-video~\cite{wang2018vid2vid} translation and re-targeting~\cite{siarohin2018animating}.

An exciting feature of \Glspl{gan} are the ability to transfer visual media across different domains. Thus, various semi-supervised and domain-adaptation tasks adopted \Glspl{gan}~\cite{ding2018one, hoffman2017cycada, shrivastava2017learning, yang20183d}. Many have leveraged synthetic data to improve model performance on real data. For example, a \gls{gan} transferred images of human eyes from the real domain to bootstrap training data~\cite{shrivastava2017learning}. Other researchers used them to synthetically generate photo-realistic images of outdoor scenes, which also aided in bettering performance in image segmentation~\cite{hoffman2017cycada}. Sometimes, labeling images captured in a controlled setting is manageable (\ie versus an uncontrolled setting). For instance, 2D body pose annotations were available \textit{in-the-wild}, while 3D annotations mostly were for images captured in a lab setting. Therefore, images with 3D annotations were used in adversarial training to predict 3D human body poses as seen \textit{in-the-wild}~\cite{yang20183d}. \cite{ding2018one} formulated one-shot recognition as a problem data imbalance and augmented additional samples in the form of synthetically generated face embeddings.

Our work differs from these others in several ways. Firstly, a majority, if not all, used a training objective that only accounts for the location of landmarks~\cite{honari2018improving,trigeorgis2016mnemonic,wang2018recurrentpami}, \ie no consideration for variance (\ie confidence). Thus, landmarks distributions have been assumed to be describable with a single parameter (\ie a mean). Networks trained this way yield an uncertainty about the prediction, while still providing a reasonable location estimate. To mitigate this, we explicitly parametrize the distribution of landmarks using location and scale. For this, we propose a KL-divergence based loss to train the network end-to-end. Secondly, previous works used \Glspl{gan} for domain adaptation in some fashion. In this work, we do not perform any adaptation between domains as in~\cite{hoffman2017cycada, shrivastava2017learning}, nor do we use any additional training labels as in~\cite{honari2018improving}. Specifically, we have \gls{d} do the quality assessment on the predicted heatmaps for a given image. The resulting gradients are used to improve the ability of the generator to detect landmarks. We show that both contributions improve accuracy when used separately. Then, the two contributions are combined to further boost state-of-the-art results.

\section{Method}
\label{sec:method}
Our training framework utilizes both labeled and unlabeled data during training. Shown in Fig.~\ref{fig:framework} are the high-level graphical depiction of cases where labels are available (blue arrows) and unavailable (red arrows). Notice the framework has two branches, supervised (Eq. \ref{eq:kl-loss}) and unsupervised (Eq. \ref{eq:kl-gan}), where only the supervised (blue arrow)uses labels to train. Next, are details for both branches.

\subsection{Fully Supervised Branch}
\label{sec:with-labels}
We define the joint distribution of the image $\image \in \mathbb{R}^{h \times w \times 3}$ and landmarks $\lmk \in \mathbb{R}^{K\times2}$ as $p(\image, \lmk)$, where $K$ is the total number of landmarks. The form of the distribution $p(\image, \lmk)$ is unknown; however, joint samples are available when labels are present (\ie $(\image, \lmk) \sim p(\image, \lmk)$). During training, we aim to learn a conditional distribution $q_\theta(\lmk | \image)$ modeled by a neural network with parameters $\theta$. Landmarks are then detected done by sampling $\tilde{\lmk} \sim q_\theta(\lmk|\image)$. We now omit parameters $\theta$ from notation for cleaner expressions. The parameter values are learned by maximizing the likelihood that the process described by the model did indeed produce the data that was observed, \ie trained by minimizing the following loss function w.r.t. its parameters: 

\begin{equation}
    \mathcal{L}(\theta) = \mathbb{E}_{(\image, \lmk) \sim p(\image, \lmk)} \| \tilde{\lmk} - \lmk \|_2.
    \label{eq:l2-loss}
\end{equation}
Alternatively, it is possible to train a neural network to predict normalized probability maps(\ie heatmaps): $\tilde{\hm} \sim q(\hm | \image)$, where $\hm \in \mathbb{R}^{K \times h \times w}$ and each $\hm_k \in \mathbb{R}^{h\times w}$ represents a normalized probability map for landmark $k$, where $k=1..K$. To get the pixel locations, one could perform the argmax operation over the heatmaps by setting $\tilde{\lmk} = \textrm{argmax}(\tilde{\hm})$). However, this operation is not differentiable and, therefore, unable to be trained end-to-end.

A differentiable variant of argmax (\ie $\sam$~\cite{chapelle2010gradient}) was recently used to localize landmarks~\cite{honari2018improving}. For the 1D case, the $\sam$ operation is expressed 

\begin{equation}
    \begin{aligned}
        \textrm{softargmax}(\beta\hm)
             & =  \sum_x \textrm{softmax}(\beta  \hm_x) \cdot x  \\
             & =  \sum_x \frac{e^{\beta  \hm_x}}{\sum_j e^{\beta  \hm_j}} \cdot x \\
             & =  \sum_x p(x) \cdot x = \mathbb{E}_\hm[x],
        \label{eq:softargmax-loss}
    \end{aligned}
\end{equation}

where $\hm_x$ is the predicted probability mass at location $x$, $\sum_j e^{\beta  \hm_j}$ is the normalization factor, and $\beta$ is the temperature factor controlling the predicted distribution~\cite{chapelle2010gradient}. We denote coordinate in boldface (\ie $\mathbf{x}=(x_1, x_2)$), and write 2D $\sam$ operation as $\tilde{\lmk}=\mathbb{E}_\hm[\mathbf{x}]$ with $\mathcal{L}_\mathrm{SAM}=\mathcal{L}(\theta)$.

Essentially, the $\sam$ operation is the expectation of the pixel coordinate over the selected dimension. Hence, the $\sam$-based loss assumes the underlying distribution is describable by just its mean (\ie location), regardless of how sure a prediction, the objective then is to match mean values. To avoid cases in which the trained model is uncertain about the predicted mean, while still yielding a low error, we parametrize the distribution using $\{\mu, \sigma\}$, where $\mu$ is the mean or the location and $\sigma$ is the variance or the scale, respectfully, for the selected distribution.

We want the model to be certain about the predictions (\ie a small variance or scale). We consider two parametric distributions$\textrm{Gaussian}(\mu, \sigma)$ and $\textrm{Laplace}(\mu, b)$ with $\sigma^2=\mathbb{E}_\hm[(\mathbf{x} - \mathbb{E}_\hm[\mathbf{x}])^2]$ and $b=\mathbb{E}_\hm[|\mathbf{x} - \mathbb{E}_\hm[\mathbf{x}]|]$. We define a function $\tau(\tilde{\hm})$ to compute the scale (or variance) of the predicted heatmaps $\tilde{\hm}$ using the location, where the locations are now the expectation of being a landmark in the heatmap space. Thus, $\tau(\tilde{\hm}) = \sum p(\mathbf{x}) ||\mathbf{x} - \tilde{\lmk}||^\alpha_\alpha$, where $\tilde{\lmk}=\mathbb{E}_\hm[\mathbf{x}]$, $\alpha=1$ for Laplacian, and $\alpha=2$ for Gaussian. Thus, $\tilde{\lmk}$ and $\tau(\tilde{\hm})$) are used to parameterize a Laplace (or Gaussian) distribution for the predicted landmarks $q(\hm | \image)$.

\begin{algorithm}[t]
\SetAlgoLined
\KwData{${\{ (\mathbf{d}_i^l, \mathbf{s}_i^l) \}_{i=1,...,n}}$,
        ${\{ (\mathbf{d}_i^u) \}_{i=1,...,m}}$

}
$\mathbf{\theta}_D, \mathbf{\theta}_G \leftarrow$ initialize network parameters\

\While{$t \leq T$}{
  $(\mathbf{D}_t^l, \mathbf{S}_t^l) \leftarrow$  sample mini-batch from labeled data\
  
  $(\mathbf{D}_t^u) \leftarrow$ sample mini-batch from unlabeled data\
  
  $\mathbf{H}_\mathrm{fake} \leftarrow G(\mathbf{D}_t^u)$\
  
  $\mathbf{H}_\mathrm{real} \leftarrow \omega(\mathbf{S}_t^l)$\
  
  $\mathcal{L}_\mathrm{adv} \leftarrow \log D([\mathbf{D}_t^l, \mathbf{H}_\mathrm{real}]) + \log( 1 - D([\mathbf{D}_t^u, \mathbf{H}_\mathrm{fake}])$
  
  $\mathcal{L}_\mathrm{G} \leftarrow$ compute loss using Eq.~\ref{eq:softargmax-loss} or Eq.~\ref{eq:kl-loss} 
  
  \vspace{0.2cm}
  
  \tcp{update model parameters}
  
  $\mathbf{\theta}_D \xleftarrow{+} - \nabla_{\mathbf{\theta}_D} \mathcal{L}_\mathrm{adv} $
  
  $\mathbf{\theta}_G \xleftarrow{+} - \nabla_{\mathbf{\theta}_G} ( \mathcal{L}_\mathrm{G} - \lambda \mathcal{L}_\mathrm{adv} )$
 }
 \caption{Training the proposed model.}
 \label{alg:training-kl}
 \vspace{2mm}
\end{algorithm}

Denoting the true conditional distribution of the landmarks as $p(\lmk|\image)$ we define the objective as follows:
\begin{equation}
    \begin{aligned}
        \mathcal{L}_\mathrm{KL} = \mathbb{E}_{(\image, \lmk) \sim p(\image, \lmk)} \Big [ \mathrm{D}_{KL} (q(\lmk | \image) || p(\lmk| \image)) \Big ],
        \label{eq:kl-loss}
    \end{aligned}
\end{equation}
where $\mathrm{D_{KL}}$ is the KL-divergence. We assumed a true distribution for the case of Gaussian (\ie $\textrm{Gaussian}(\mathbf{\mu} , 1)$, where $\mu$ is the ground-truth locations of the landmarks). For the case with Laplace, we sought $\textrm{Laplace}(\mathbf{\mu}, 1)$. KL-divergence conveniently has a closed-form solution for this family of exponential distributions~\cite{hoffman2013stochastic}. Alternatively, sampling yields an approximation. The blue arrow in Fig.~\ref{fig:framework} represent the labeled branch of the framework.

\begin{figure}[t!]
    \centering
    \includegraphics[width=3.2in]{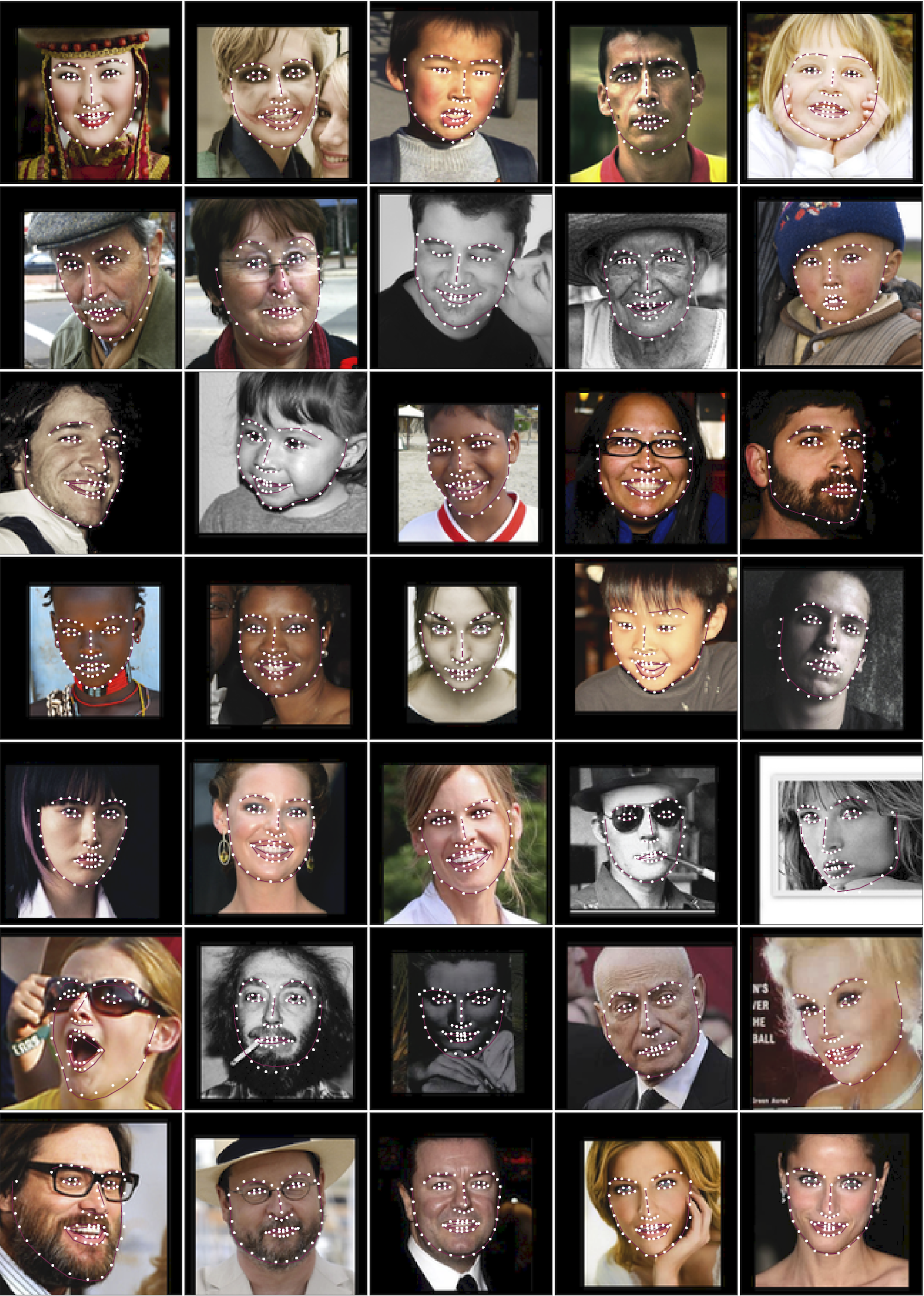}
    \caption{Random samples of landmarks predicted using $\lkl$ (white), with the ground truth drawn as line segments (\textcolor{red}{red}). Notice the predicted points tend to overlap with the ground-truth. Best viewed in color. Zoom-in for greater detail.}
    \label{fig:faces}
     \vspace{-2mm}
\end{figure}

Statistically speaking, given two estimators with different variances, we would prefer one that has a smaller variance (see \cite{domingos2000unified} for an analysis of the bias-variance trade-off). A lower variance implies higher confidence in the prediction. To this end, we found an objective measuring distance between distributions is accurate and robust. The neural network must satisfy an extra constraint on variance and, thus, yields predictions of higher certainty. See higher confident heatmaps in Fig.~\ref{fig:kl-versus-softargmax} and Fig.~\ref{fig:qualitative-figure}. The experimental evaluation further validates this (Table \ref{tab:face-comparisons} and Table \ref{tab:model-sizes}). Also, Fig.~\ref{fig:faces} shows sample results.

\subsection{Unsupervised Branch}\glsreset{d}\glsreset{g}
\label{sec:no-labels}
The previous section discusses several objectives to train the neural network with the available paired or fully labeled data (\ie $(\image^l, \lmk^l)$). We denote data samples with the superscript $l$ to distinguish them from unpaired or unlabeled data $(\image^u)$. In general, it is difficult for a human to label many images with landmarks. Hence, unlabeled data are abundant and easier to obtain, which calls for capitalizing on this abundant data to improve training. In order to do so, we adapt the adversarial learning framework for landmark localization. We treat our landmarks predicting network as a \gls{g}, $G=q(\hm | \image)$. \gls{d} takes the form $D([ \image, \hm ])$, where $[\cdot, \cdot]$ is a tensor concatenation operation . We define the real samples for \gls{d} as $\{\image^l, \hm= \omega(\lmk^l)\}$, where $\omega(\cdot)$ generates the true heatmaps given the locations of the ground-truth landmarks. Fake samples are given by $\{ \image^u, \tilde{\hm} \sim q(\hm | \image^u) \}$. With this notation, and we define the min-max objective for landmark localization as:

\begin{eqnarray}
    \min_{G} \max_{D} \mathcal{L}_\mathrm{adv}(D,G),
    \label{eq::gan-problem}
\end{eqnarray}

where $\mathcal{L}_\mathrm{adv}(D,G)$ writes as:

\begin{eqnarray}
        \!\!\!&  &\!\!\! \mathbb{E}_{(\image^l, \lmk^l) \sim p(\image, \lmk)} \Big [\log D([\image^l, \omega(\lmk^l)]) \Big ] + \nonumber\\
        \!\!\!&   &\!\!\!\mathbb{E}_{(\image^u) \sim p(\image)} \Big [\log( 1 - D([\image^u, G(\image^u))]) \Big ].
\end{eqnarray}

In this setting, provided an input image, the goal of \gls{d} is to learn to decipher between the \textit{real} and \textit{fake} heatmaps from appearance. Thus, the goal of \gls{g} is to produce \textit{fake} heatmaps that closely resemble the \textit{real}. Within this framework, \gls{d} intends to provide additional guidance for \gls{g} by learning from both labeled and unlabeled data. The objective in Eq.~\ref{eq::gan-problem} is solved using alternating updates.

\subsection{Training}
\label{sec:training}
We fused the $\sam$-based and adversarial losses as 
\begin{eqnarray}
       \min_G \Big ( \max_D \big ( 
                \lambda \cdot \mathcal{L}_\mathrm{adv}(G, D)
            \big )
            + \mathcal{L}_\mathrm{SAM}(G)
        \Big ),
\label{eq:sam-gan}
\end{eqnarray}
with the KL-divergence version of the objective defined as:
\begin{eqnarray}
       \min_G \Big ( \max_D \big ( 
                \lambda \cdot \mathcal{L}_\mathrm{adv}(G, D)
            \big )
            + \mathcal{L}_\mathrm{KL}(G)
        \Big ),
\label{eq:kl-gan}
\end{eqnarray}
with the weight for the adversarial loss $\lambda=0.001$. This training objective includes both labeled and unlabeled data in the formulation. In the experiments, we show that this combination significantly improves the accuracy of our approach. We also argue that the $\sam$-based version cannot fully utilize the unlabeled data since the predicted heatmaps differ too much from the \textit{real} heatmaps. See Algorithm~\ref{alg:training-kl} for the training procedure for $T$ steps of the proposed model. We show the unlabeled branch of the framework is shown graphically in red arrows (Fig.~\ref{fig:framework}).

\begin{table}\centering
	\caption {\glsreset{g}Architecture of the \gls{g}. Layers listed with the size and number of filters (\ie $h$ $\times$ $w$ $\times$ $n$). DROP, MAX, and UP stand for dropout (probability 0.2), max-pooling (stride 2), and bilinear upsampling (2$x$). Note the skip connections about the bottleneck: coarse-to-fine, connecting encoder (\ie $E_{ID}$) to the decoder (\ie $D_{ID}$) by concatenating feature channels before fusion via fully-connected layers. Thus, all but the 2 topmost layers had feature dimensions and the number of feature maps preserved (\ie layers that transformed feature maps to $K$ heatmaps). A stride of 1 and padded such to produce output sizes listed.}

\resizebox{\linewidth}{!}{
    \begin{tabularx}{290pt}{lll}\toprule
		{} & \textbf{Layers} & \textbf{Tensor Size}\\\midrule
	    Input          & RGB image, no data augmentation                & 80 x 80 x 3 \\
		Conv($E_1$)    & 3 $\times$ 3 $\times$ 64, LReLU, DROP, MAX                   & 40 $\times$ 40 $\times$ 64\\
        Conv($E_2$)    & 3 $\times$ 3 $\times$ 64, LReLU, DROP, MAX                   & 20 $\times$ 20 $\times$ 64\\
        Conv($E_3$)    & 3 $\times$ 3 $\times$ 64, LReLU, DROP, MAX                   & 10 $\times$ 10 $\times$ 64\\
        Conv($E_4$)    & 3 $\times$ 3 $\times$ 64, LReLU, DROP, MAX                   & 5 $\times$ 5 $\times$ 64\\
         
        Conv($D_4$)    & 1 $\times$ 1 $\times$ 64 $+E_4$, LReLU, DROP, UP      & 10 $\times$ 10 $\times$ 128\\
         
        Conv($D_F$)    & 5 $\times$ 5 $\times$ 128, LReLU                             & 20 $\times$ 20 $\times$ 128\\
        Conv($D_3$     & 1 $\times$ 1 $\times$ 64 $+E_3$, LReLU, DROP, UP       & 20 $\times$ 20 $\times$ 128\\
         
        Conv($D_F$)    & 5 $\times$ 5 $\times$ 128, LReLU, DROP                       & 40 $\times$ 40 $\times$ 128\\
        Conv($D_2$)    & 1 $\times$ 1 $\times$ 64 $+E_2$, LReLU, DROP,  UP      & 40 $\times$ 40 $\times$ 128\\
         
        Conv($D_F$)    & 5 $\times$ 5 $\times$ 128, LReLU, DROP                       & 80 $\times$ 80 $\times$ 128 \\
        Conv($D_1$)    & 1 $\times$ 1 $\times$ 64 $+E_1$, LReLU, DROP, UP     & 80 $\times$ 80 $\times$ 128\\
        
        Conv($D_F$)    & 5 $\times$ 5 $\times$ 128, LReLU, DROP                       & 80 $\times$ 80 $\times$ 128\\
        Conv($D_F$)    & 1 $\times$ 1 $\times$ 68, LReLU, DROP                        & 80 $\times$ 80 $\times$ 68\\
        Output         & 1 $\times$ 1 $\times$ 68                                     & 80 $\times$ 80 $\times$ 68\\
\bottomrule
\end{tabularx}
}
\label{tbl:arch}
\end{table}

\subsection{Implementation}
\label{sec:implementation}
We follow the ReCombinator network (RCN) initially proposed in~\cite{honari2016recombinator}. Specifically, we use a 4-branch RCN as our base model, with input images and output heatmaps of size 80$\times$80. Convolutional layers of the encoder consist of 64 channels, while the convolutional layers of the decoder output 64 channels out of the 128 channels at its input (\ie 64 channels from the previous layer concatenated with the 64 channels skipped over the bottleneck via branching). We applied Leaky-ReLU, with a negative slope of 0.2, on all but the last convolution layer. See Table~\ref{tbl:arch} for details on the generator architecture. Drop-out followed this, after all but the first and last activation. We use Adam optimizer with a learning rate of 0.001 and weight decay of $10^{-5}$. In all cases, networks were trained from scratch, using no data augmentation nor any other 'training tricks.' 

\gls{d} was a 4-layered PatchGAN~\cite{isola2017image}. Before each convolution layer Gaussian noise ($\sigma=0.2$) was added~\cite{tulyakov2017mocogan}, and then batch-normalization (all but the top and bottom layers) and Leaky-ReLU with a negative slope of 0.2 (all but the top layer). The original RGB image was stacked on top of the $K$ heatmaps from \gls{g} and fed as the input of \gls{d} (Fig.~\ref{fig:framework}). Thus, \gls{d} takes in ($K$ + 3) channels. We set $\beta=1$ for \ref{eq:softargmax-loss}. Pytorch was used to implement the entire framework. An important note to make is that models optimized with Laplace distribution consistently outperformed the Gaussian-based. For instance, our $\lkl$ baseline had a \gls{nmse} of 4.01 on 300W, while Gaussian-based got 4.71. Thus, the sharper,``peakier'' Laplace distribution proved to be more numerically stable under current network configuration, as Gaussian required a learning rate a magnitude smaller to avoid vanishing gradients. Indeed, we used Laplace.

\begin{table}[t!]
\centering
\caption{\gls{nmse} on \gls{aflw} and 300W normalized by the square root of BB area and interocular distance, respectfully.}
\label{tab:face-comparisons}
\centering
\resizebox{\linewidth}{!}{
    \begin{tabularx}{300pt}{l@{\hskip .25in}c@{\hskip .4in}ccc}\toprule
                                      & \multirow{2}{*}{\textbf{\gls{aflw}}}       & \multicolumn{3}{c}{\textbf{300W}}        \tabularnewline
                                      &                             & \textbf{Common}        & \textbf{Challenge}   & \textbf{Full}              \\\midrule
                                      
SDM~\cite{xiong2013supervised}            & 5.43                        & 5.57          & 15.40         & 7.52              \tabularnewline
LBF~\cite{ren2014face}                    & 4.25                        & 4.95          & 11.98         & 6.32         		\tabularnewline
MDM~\cite{trigeorgis2016mnemonic}         & -                           & 4.83          & 10.14         & 5.88         		\tabularnewline
TCDCN~\cite{zhang2014facial}                & -                           & 4.80          & 8.60          & 5.54        		\tabularnewline
CFSS~\cite{zhu2015face}                    & 3.92                        & 4.73          & 9.98          & 5.76         		\tabularnewline
CFSS~\cite{lv2017deep}                     & 2.17                        & 4.36          & 7.56          & 4.99         		\tabularnewline
RCSR~\cite{wang2018recurrentpami}          & -                           & 4.01          & 8.58          & 4.90         		\tabularnewline
RCN+ (L$+$ELT)~\cite{honari2018improving}            & \textbf{1.59}               & 4.20          & 7.78          & 4.90         		\tabularnewline
CPM $+$ SBR~\cite{dong2018supervision}            & 2.14                        & 3.28          & 7.58          & 4.10     	    	\tabularnewline\midrule
$\Sam$                                & 2.26                        & 3.48          & 7.39          & 4.25    		    \tabularnewline
$\Sam$+D(10K)                         & -                           & 3.34          & 7.90          & 4.23     		    \tabularnewline
$\Sam$+D(30K)                         & -                           & 3.41          & 7.99          & 4.31     		    \tabularnewline
$\Sam$+D(50K)                         & -                           & 3.41          & 8.06          & 4.32     		    \tabularnewline
$\Sam$+D(70K)                         & -                           & 3.34          & 8.17          &  4.29     		    \tabularnewline\midrule
$\lkl$                                & 1.97                        & 3.28          & 7.01          & 4.01     			\tabularnewline
$\lkl$+D(10K)                         & -                           & 3.26          & 6.96          & 3.99            \tabularnewline
$\lkl$+D(30K)                         & -                           & 3.29          & 6.74          & 3.96              \tabularnewline
$\lkl$+D(50K)                         & -                           & 3.26          & \textbf{6.71} & 3.94              \tabularnewline
$\lkl$+D(70K)                         & -                           & \textbf{3.19} & 6.87          & \textbf{3.91}     \tabularnewline\bottomrule
    \end{tabularx}
  }
\end{table}

\section{Experiments}

\label{sec:experiments}
We evaluated the proposed on two widely used benchmark datasets for face alignment. No data augmentation techniques used when training our models nor was the learning rate dropped: this leaves no ambiguity into whether or not the improved performance came from training tricks or the learning component itself. All results for the proposed were from models trained for 200 epochs.

We next discuss the metric used to evaluate performance, \gls{nmse}, with differences between datasets in the normalization factor. Then, the experimental settings, results, and analysis for each dataset are covered separately. Finally, ablation studies show characterizations of critical hyper-parameters and, furthermore, the robustness of the proposed $\lkl$+D(70K) with a comparable performance with just $1/8$ the number of feature channels and $>$20 fps.

\subsection{Metric}
Per convention~\cite{bulat2017far, cristinacce2006feature, sagonas2013300}, \gls{nmse}, a normalized average of euclidean distances, was used. Mathematically speaking: 

\begin{equation}
    \text{NMSE} = \sum_{k=1}^{K} \frac{\|s_k - \tilde{s}_k\|_2}{K\times d}, 
    \label{eqn:NMSE}
\end{equation}
where the number of visible landmarks set as $K$, $k=\{1,2,...,K\}$ are the indices of the visible landmark, the normalization factor $d$ depends on the face size, and $s_k \in \mathbb{R}^2$ and $\tilde{s}_k \in \mathbb{R}^2$ are the ground-truth and predicted coordinates, respectfully. The face size $d$ ensured that the \gls{nmse} scores across faces of different size were fairly weighted. Following predecessors, \gls{nmse} was used to evaluate both datasets, except with different points referenced to calculate $d$. The following subsections provide details for finding $d$.

\subsection{300W + MegaFace}
The 300W dataset is amongst the most popular datasets for face alignment. It has 68 visible landmarks (\ie $K=68$) for 3,837 images (\ie 3,148 training and 689 test). We followed the protocol of the 300W challenge~\cite{sagonas2013300} and evaluated using \gls{nmse} (Eq.~\ref{eqn:NMSE}), where $d$ is set as the inter-ocular distance (\ie distance between outer corners of the eyes). Per convention, we evaluated different subsets of 300W (\ie \textit{common} and \textit{challenge}, which together form \textit{full}).

We compared the performance of the proposed objective trained in a semi-supervised fashion. During training, 300W dataset made-up the labeled data (\ie \textit{real}), and a random selection from MegaFace provided the unlabeled data (\ie \textit{fake})~\cite{nech2017level}. MTCNN\footnote{\url{https://github.com/davidsandberg/facenet}} was used to detect five landmarks (\ie eye pupils, corners of the mouth, and middle of nose and chin)~\cite{zhang2016joint}, which allowed for similar face crops from either dataset. Specifically, we extended the square hull that enclosed the five landmarks by 2$\times$ the radii in each direction. In other words, the smallest bounding box spanning the 5 points (\ie the outermost points lied on the parameter), and then transformed from rectangles-to-squares with sides of length 2$\times\max(height, width)$. Note that the midpoint of the original rectangle was held constant to avoid shift translations (\ie rounded up a pixel if the radius was even and extended in all directions).

\begin{figure}
    \begin{tabular}{p{.28in}p{.27in}p{.75in}p{.3in}p{.27in}p{.3in}p{.1in}}
        &\scriptsize\rotatebox{15}{L-KL+D(70K)}&\scriptsize\rotatebox{15}{SAM+D(80K)}&\scriptsize\rotatebox{15}{L-KL+D(70K)}&\scriptsize\rotatebox{15}{SAM+D(80K)}&  \\
    \multicolumn{7}{c}{
    \includegraphics[width=\linewidth, trim={1.75mm 0mm 0.0mm 0mm},clip]{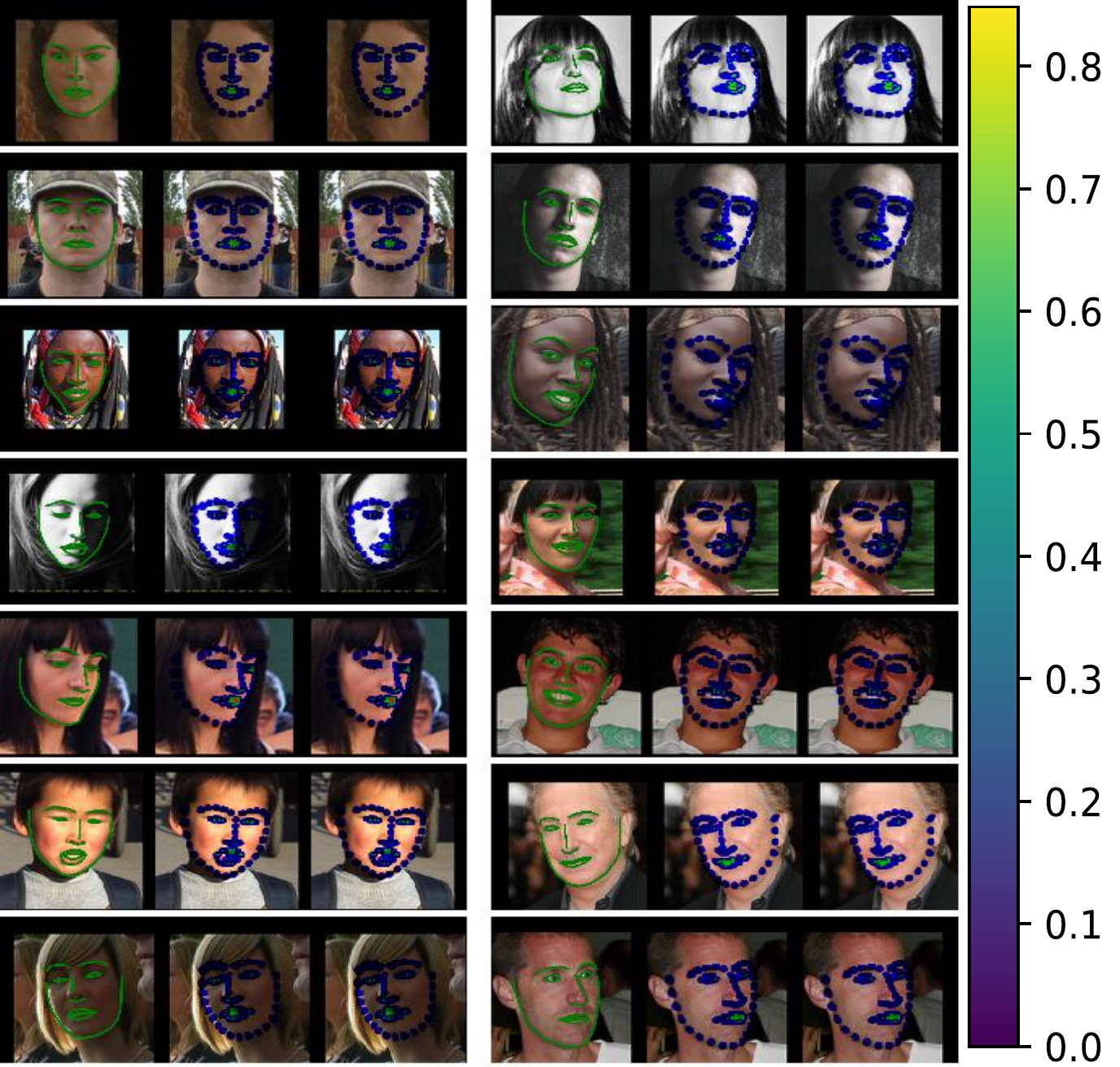}
    }
    \end{tabular}
    \caption{Random samples (300W). Heatmaps predicted by our $\lkl$+D(70K) (middle, \ie L-KL+D(70K)) and $\sam$+D(70K) (right, \ie SAM+D(70K)) alongside face images with ground-truth sketched on the face (left). For this, colors were set by value for the $K$ heatmaps generated for each landmark (\ie range of [0, 1] as shown in color bar), and then were superimposed on the original face. Note that the KL-divergence loss yields predictions of much greater confidence and, hence, produced separated landmarks when visualized heatmap space. In other words, the proposed has minimal spread about the mean, as opposed to the $\sam$-based model with heatmaps with individual landmarks smudged together. Best viewed electronically. }\label{fig:qualitative-figure}
\end{figure}
The $\lkl$+D(70K) model obtained state-of-the-art on 300W, yielding the lowest error on 300W (Table~\ref{tab:face-comparisons}~(300W columns)). $\lkl$+D($N$) and $\sam$+D($N$) denote the models trained with unlabeled data, where $N$ representing the number of unlabeled images added from MegaFace.

First, notice that $\lkl$ trained without unlabeled data still achieved state-of-the-art. The $\lkl$-based models then showed relative improvements with more unlabeled data added. The $\sam$-based models cannot fully take advantage of the unlabeled data without minimizing for variance (\ie generates heatmaps of less confidence and, thus, more spread). Our $\lkl$, on the other hand, penalizes for spread (\ie scale), making the job of \gls{d} more challenging. As such, $\lkl$-based models benefit from increasing amounts of unlabeled data.

Also, notice the largest gap between the baseline models~\cite{dong2018supervision} and our best $\lkl$+D(70K) model on the different sets of 300W. Adding more unlabeled helps more (\ie $\lkl$ vs. $\lkl$+D(70K) improvement is about 2.53\%). However, it is essential to use samples not covered in the labeled set. To demonstrate this, we set the \textit{real} and \textit{fake} sets to 300W (\ie $\image^l = \image^u$ in the second term of Eq.~\ref{eq:kl-gan}). \gls{nmse} results for this experiment are listed as follows: $\lkl$+D(300W) 4.06 (baseline-- 4.01) and $\sam$+D(300W) 4.26 (baseline-- 4.24). As hypothesized, all the information from the labeled set had already been extracted in the supervised branch, leaving no benefit of using the same set in the unsupervised branch. Therefore, more unlabeled data yields more hard negatives to train with, which improves the accuracy of the rarely seen samples (Table~\ref{tab:face-comparisons}~(300W \textit{challenge} set)). Our best model was $\approx$2.7\% better than~\cite{dong2018supervision} on easier samples (\ie \textit{common}), $\approx$4.7\% better on average (\ie \textit{full}), and, moreover, $\approx$9.8\% better on the more difficult (\ie \textit{challenge}), $\approx$4.7\% better on average (full), and, moreover, $\approx$9.8\% better on the more difficult (challenge). These results further highlight the advantages of training with the proposed $\lkl$ loss, along with the adversarial training framework.

Additionally, the adversarial framework further boosted our 300W baseline was further boosted by (\ie more unlabeled data yields a lower \gls{nmse}). Specifically, we demonstrated this by pushing state-of-the-art of the proposed on 300W from a \gls{nmse} of 4.01 to 3.91 (\ie no unlabeled data to 70K unlabeled pairs, respectfully). There were boosts at each step size of \textit{full} (\ie larger $N$ $\rightarrow$ \gls{nmse}).

We randomly selected unlabeled samples for $\lkl$+D(70K) and $\sam$+D(70K) to visualize predicted heatmaps (Fig.~\ref{fig:qualitative-figure}). In each case, the heatmaps produced by the softargmax-based models spread wider, explaining the worsened quantitative scores (Table~\ref{tab:face-comparisons}). The models trained with the proposed contributions tend to yield higher probable pixel location (\ie a more concentrated predicted heatmaps). For most images, the heatmaps generated by models trained with the $\lkl$ loss have distributions for landmarks that were more confident and properly distributed: our $\lkl$+D(70K) yielded heatmaps that vary 1.02 pixels from the mean, while $\sam$+D(70K) has a variation of 2.59. Learning the landmark distributions with our $\lkl$ loss is conceptually and theoretically intuitive (Fig.~\ref{fig:kl-versus-softargmax}). Moreover, it is experimentally proven (Table~\ref{tab:face-comparisons}).

\subsection{The \gls{aflw} dataset}
We evaluated the $\lkl$ loss on the \gls{aflw} dataset~\cite{koestinger2011annotated}. \gls{aflw} contains 24,386 faces with up to 21 landmarks annotations and 3D head pose labels. Following~\cite{honari2018improving}, 20,000 faces were used for training with the other 4,386 for testing. We ignored the two landmarks for the left and right earlobes, leaving up to 19 landmarks per face~\cite{dong2018supervision}.

Since faces of \gls{aflw} have such variety head poses, most faces have landmarks out of view (\ie missing). Thus, most samples were not annotated with the complete 19 landmarks, meaning that it does not allow for a constant sized tensor (\ie \textit{real} heatmaps) for the adversarial training. Therefore, we compared the $\sam$ and KL-based objectives with existing state-of-the-art. The face size $d$ for the \gls{nmse} was the square root of the bounding box hull~\cite{bulat2017far}.

Our $\lkl$-based model scored results comparable to existing state-of-the-art (\ie RCN+ (L$+$ELT)~\cite{honari2018improving}) on the larger, more challenging \gls{aflw} dataset while outperforming all others. It is essential to highlight here that ~\cite{honari2018improving} puts great emphasis on data augmentation, while we do not apply any. Also, since landmarks are missing in some samples (\ie no common reference points exist across all samples), we were unable to prepare faces for our semi-supervised component-- a subject for future work.

\begin{table}[t!]
\centering
\caption{\gls{nmse} on 300W (full set) for networks trained with fewer channels in each convolutional layer by $1/16$, $1/8$, $1/4$, $1/2$, and unmodified in size (\ie the original) listed from left-to-right. We measured performance with a 2.8GHz Intel Core i7 CPU.}
\label{tab:model-sizes}
\centering
\resizebox{\linewidth}{!}{
    \begin{tabularx}{300pt}{lccccc}\toprule

            & \multicolumn{5}{c}{Number of parameters, millions}\vspace{.2mm}\tabularnewline 
            \cline{2-6}
                                      &      $0.0174$                        & $0.0389$ 
                                         & $0.1281$
                                         & $0.4781$   & $1.8724$
                                      \tabularnewline\midrule
$\Sam$                                &         9.79               &             6.86 &  4.83      & 4.35 & 4.25     		    \tabularnewline
$\Sam$+D(70K)                         &    9.02                        &           6.84  & 4.85  &   4.38  & 4.29      	\tabularnewline	    
$\lkl$                                &          7.38            &      5.09    &  4.39    & 4.04 & 4.01     			\tabularnewline
$\lkl$+D(70K)                         &                    \textbf{7.01}       & \textbf{4.85} &       \textbf{4.30}  &  \textbf{3.98} & \textbf{3.91}   
\tabularnewline\midrule\midrule
Storage (MB) &0.076 & 0.162 & 0.507& 1.919&7.496 \tabularnewline
Speed (fps) & 26.51 &21.38 &16.77& 11.92 & 4.92
\tabularnewline\bottomrule
    \end{tabularx}
    }
    \vspace{-3mm}
\end{table}

\begin{figure}[t!]
    \centering
    \includegraphics[width=0.5\linewidth, trim={1.0mm 0mm .5mm 0mm},clip]{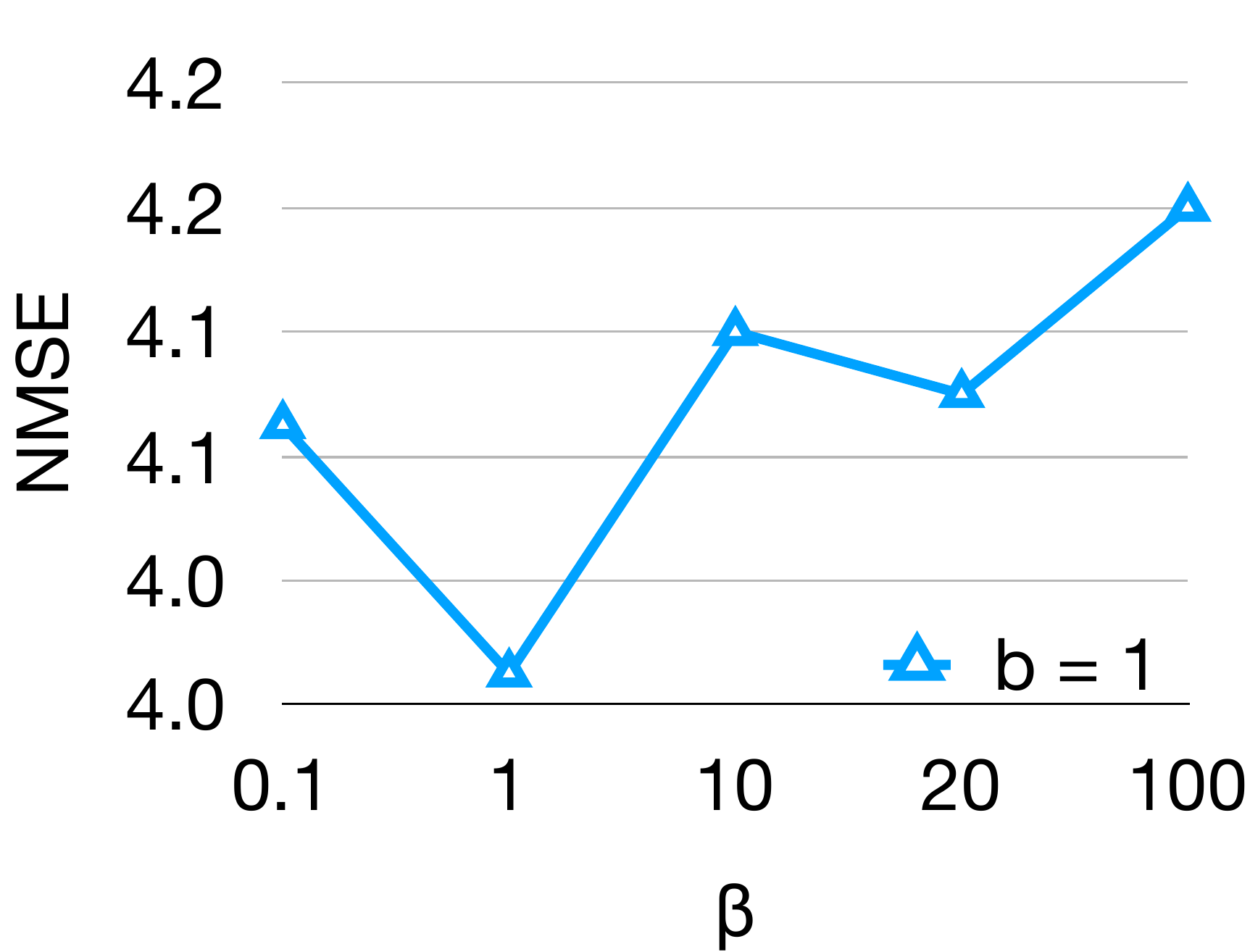}
    \includegraphics[width=0.48\linewidth, trim={10.0mm 0mm 0.5mm 0mm},clip]{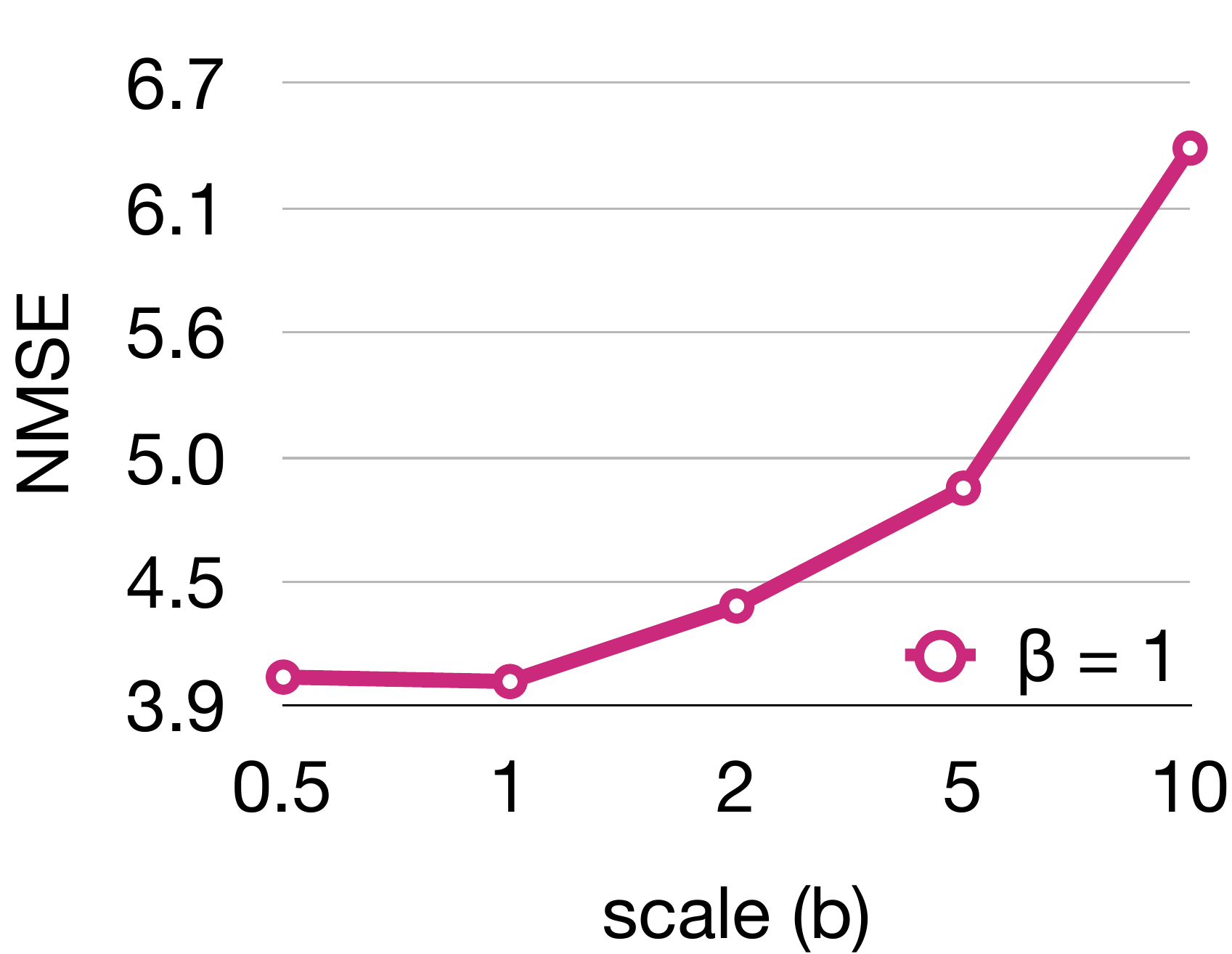}

    \caption{Results of ablation study on $\lkl$.}
    \label{fig:ablation}
\end{figure}

\subsection{Ablation Study}
The error is next measured as a function of model size (Table~\ref{tab:model-sizes}), along with different$\beta$ values (Eq. \ref{eq:softargmax-loss}) and scales $b$ used to parameterize the Laplacian (Fig.~\ref{fig:ablation}). The latter characterizes the baseline and supports the values used for these hyper-parameters, while the former reveals a critical characteristic for the practicality of the proposed. 

Specifically, we decreased the model size by reducing the number of channels at each convolutional layer by factors of 2. The $\sam$-based model worsened by about 47\% and 79\% in \gls{nmse} at an and the channel count, respectfully (\ie 4.25 $\rightarrow$ 6.86 and 9.79). $\lkl$, on the other hand, decreased by about 24\% with an $8^{th}$ and 59\% with a $16^{th}$ the number of channels (\ie 4.01 $\rightarrow$ 5.09 and 7.38, respectfully). Our model trained with unlabeled data (\ie $\lkl$+D(70K)) dropped just about 21\% and 57\% at factors of 8 and 16, respectfully (\ie 3.91 $\rightarrow$ 4.85 and 7.01). In the end, $\lkl$+D(70K) proved best with reduced sizes: with $<$0.040M parameters, it still compares to previous state-of-the-art~\cite{honari2018improving,lv2017deep,wang2018recurrentpami}, which is a clear advantage. For instance, SDM~\cite{xiong2013supervised}, requires 1.693M parameters (25.17MB) for 7.52 in \gls{nmse} (300W \textit{full}).\footnote{\url{https://github.com/tntrung/sdm\_face\_alignment}} Yet our smallest and next-to-smallest get 7.01 and 4.85 with only 0.174M (0.076 MB) and 0.340M (0.166 MB) parameters.

The processing speed also boosts with fewer channels (\ie to train and at inference). For instance, the model reduced by a factor of 16 processes 26.51 frames per second (fps) on a CPU of Macbook Pro (\ie 2.8GHz Intel Core i7), with the original running at 4.92 fps. Our best $\lkl$-based model proved robust to size reduction, obtaining 4.85 \gls{nmse} at 21.38 fps when reduced by $1/8$. 

\section{Conclusions}
\label{sec:conclusion}
We demonstrated the benefits of the proposed $\lkl$ loss and leveraging unlabeled data in an adversarial training framework. Hypothetically and empirically, we showed the importance of penalizing a landmark predictor's uncertainty. Thus, training with the proposed objective yields predictions of higher confidence, outperforming previous state-of-the-art methods. We also revealed the benefits of adding unlabeled training data to boost performance via adversarial training. In the end, our model performs state-of-the-art on all three splits of the renown 300W (\ie \textit{common}, \textit{challenge}, and \textit{full}), and second-to-best on the \gls{aflw} benchmark. Also, we demonstrate the robustness of the proposed by significantly reducing the number of parameters. Specifically, with $1/8$ the number of channels (\ie $<$170Kb on disk), the proposed still yields an accuracy comparable to the previous state-of-the-art in real-time (\ie 21.38 fps). Thus, the contributions of the proposed framework are instrumental for models intended for use in real-world production.

{\small
\bibliographystyle{ieee_fullname}
\balance

\bibliography{egbib}
}

\end{document}